\documentclass[acmtog,nonacm]{acmart}
\AtBeginDocument{%
  }

\usepackage{algorithm}
\usepackage{cleveref}

\DeclareMathOperator{\Var}{Var}
\newcommand{\E}{\mathbb{E}}
\newcommand{\proj}{\mathcal{P}}
\setcopyright{none}
\renewcommand\footnotetextcopyrightpermission[1]{}

\hypersetup{
  pdfpublisher={},
  pdfcopyright={},
  pdflicenseurl={},
  pdfsubject={}
}

\begin{document}

\title{Bridging Rendering and Generative Modeling with Monte Carlo Transport Scheduling}

\author{Junwei Shu}
\orcid{0009-0006-1197-032X}
\email{51265901091@stu.ecnu.edu.cn}
\affiliation{%
  \institution{East China Normal University}
  \city{Shanghai}
  \country{China}
}

\author{Wenjie Liu}
\orcid{0009-0008-0088-7701}
\email{51265901068@stu.ecnu.edu.cn}
\affiliation{%
  \institution{East China Normal University}
  \city{Shanghai}
  \country{China}
}

\author{Hantang Liu}
\orcid{0000-0003-2844-2436}
\email{defmacro1991@gmail.com}
\affiliation{%
  \institution{KunByte}
  \city{Shanghai}
  \country{China}
}

\author{Changbo Wang}
\authornote{Corresponding authors.}
\orcid{0000-0001-8940-6418}
\email{cbwang@dase.ecnu.edu.cn}
\affiliation{%
  \institution{East China Normal University}
  \city{Shanghai}
  \country{China}
}

\author{Yang Li}
\authornotemark[1]
\orcid{0000-0001-9427-7665}
\email{yli@cs.ecnu.edu.cn}
\affiliation{%
  \institution{East China Normal University}
  \city{Shanghai}
  \country{China}
}

\renewcommand{\shortauthors}{Shu et al.}

\begin{abstract}
  Monte Carlo rendering and modern generative models both transform uncertain states into structured images, yet they are usually studied as separate processes. We introduce Monte Carlo Transport Scheduling, a framework that treats progressive path tracing as a continuous sampling-driven transport process. Our key observation is that the renderer already produces physically valid states along this process: nested Monte Carlo estimates trace a refinement trajectory whose natural time coordinate follows from sampling variance. This view leads to a continuous training framework that learns from real render endpoints rather than synthetic interpolants, preserving the statistical structure of Monte Carlo estimation while enabling arbitrary-step neural refinement. We evaluate the framework on a controlled rendering benchmark designed to separate transport difficulty from scene context, and show that it yields stable render refinement, supports continuous stopping between rendering states, and transfers as a physical prior for frozen generative samplers. These results suggest a common continuous-time substrate for rendering and generation, where Monte Carlo sampling provides both the physical states and the supervision for learning image transport.
\end{abstract}

\begin{teaserfigure}
  \includegraphics[width=\textwidth]{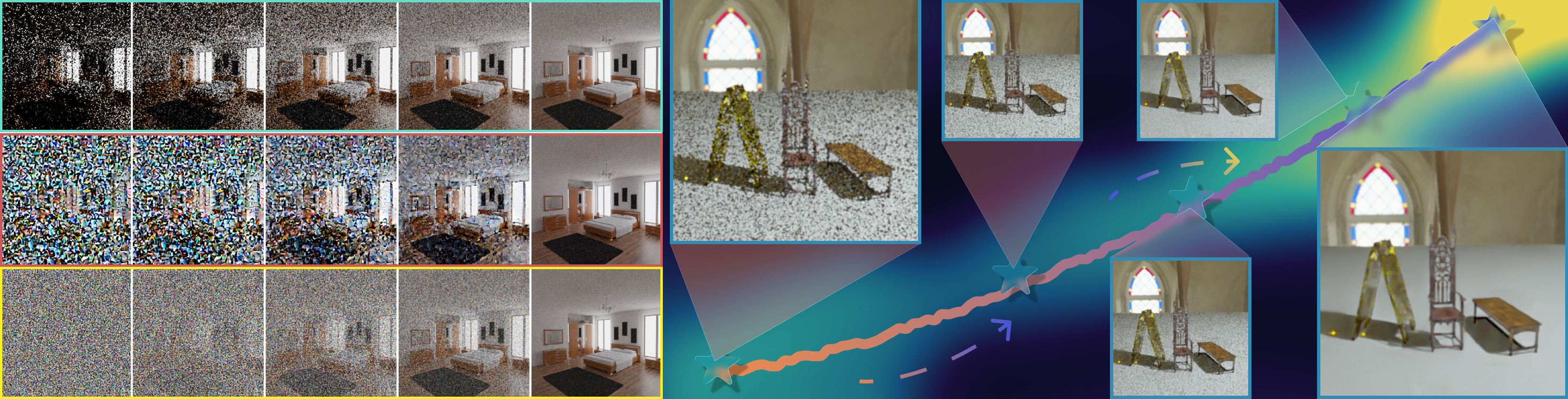}
  \caption{Monte Carlo Transport Scheduling links rendering and generation through
a shared noisy-to-clean transport view. Left: progressive path tracing,
diffusion denoising, and Rectified Flow sampling each trace a trajectory from
uncertainty to structure. Right: path tracing additionally provides real nested
Monte Carlo endpoints along this process; we use these endpoints to define a
continuous render-state space and learn a transport field toward the clean-limit
image.}
  \Description{Teaser visualization of a Monte Carlo render-state space. A
  low-sample input is connected to generated intermediate and clean image states
  by abstract trajectories representing real Monte Carlo refinement and the
  learned transport rollout.}
  \label{fig:teaser}
\end{teaserfigure}

\maketitle

\section{Introduction}

Diffusion-based generative models~\citep{Song2021ScoreSDE,Ho2020DDPM,Karras:2022:EDS} have become the dominant paradigm for high-fidelity image synthesis, learning to reverse a noise-adding process that transforms Gaussian noise into structured visual content.
Despite their empirical success, these models remain detached from the physical principles of image formation: the denoising dynamics are data-driven, and the intermediate states along the trajectory carry no physical meaning beyond the hand-crafted interpolation schedule.

In contrast, Monte Carlo path tracing~\citep{Kajiya:1986:RTE} produces images through a physically grounded sampling process.
Each additional sample reduces the estimator's variance; the rendered image, noisy at 1~sample per pixel, sharpens continuously toward the ground truth.
This progressive refinement follows a single physical convergence law---the Central Limit Theorem---and forms a natural trajectory whose intermediate states are real Monte Carlo estimates rather than synthetic interpolants.
Strikingly, this trajectory from noise to clarity mirrors the structure of diffusion-based generation, yet the connection has remained unexploited: diffusion models treat denoising as a learned data-driven schedule, while rendering treats sampling as open-ended numerical integration, and neither field has formulated a shared time axis on which both processes can operate.

In this paper, we introduce \emph{Monte Carlo Transport Scheduling}, a framework that derives a continuous time axis directly from the convergence properties of Monte Carlo integration, and uses this axis to train neural solvers that work across both rendering and generative modeling.

The core object is the \emph{variance time} $\tau(n)$, a schedule mapping any sample count $n$ to a normalized coordinate in $[0,1]$, derived from the $1/\sqrt{n}$ convergence law of the Central Limit Theorem.
This schedule turns discrete SPP ladders into continuous refinement trajectories whose physical validity is preserved at every endpoint, because training uses only real Monte Carlo states---never synthetic interpolants.

The schedule admits two natural parameterizations.
In a Rectified Flow framework, it defines an \emph{MC velocity} target: a field that transports any real render state directly toward the clean limit, enabling arbitrary-step refinement with minimal quality degradation.
In a diffusion framework, it defines an \emph{MC noise} target: a denoising objective that aligns the Monte Carlo variance decay with the diffusion reverse process, allowing the schedule to be injected into pretrained DDPM samplers as a physical rendering-noise prior.
Both parameterizations share the same variance-time axis, the same endpoint-only training constraint, and the same transport interpretation.

We evaluate Monte Carlo Transport Scheduling through three lenses: render refinement, where the solver matches or exceeds traditional baselines while retaining stability from 1 to 50 steps; bridge continuity, where the learned field stops near real non-dyadic Monte Carlo endpoints outside the training ladder; and generative trajectory injection, where the transport schedule transfers as a physical prior for frozen diffusion and RF samplers, improving no-reference image quality metrics.
All experiments are conducted on a controlled rendering benchmark with balanced transport regimes, scene contexts, nested-prefix SPP ladders, and strict scene-content-disjoint splits.
In summary, our contributions are:
\begin{itemize}
  \item \textbf{Monte Carlo Transport Scheduling}: a continuous-time framework that derives a variance-time schedule from the convergence law of Monte Carlo integration, embedding discrete SPP ladders into a domain compatible with both Rectified Flows and diffusion models.

  \item \textbf{Endpoint-anchored training for transport solvers}: a formulation that optimizes velocity or denoising fields exclusively on real Monte Carlo endpoints. MC velocity and MC noise targets---corresponding to flow-based and diffusion-based parameterizations of the same schedule---achieve stable arbitrary-NFE inference and transfer as a physical prior for frozen generative samplers.

  \item \textbf{Controlled benchmark and evidence of transport reuse}: a multi-axis rendering benchmark with nested-prefix SPP ladders, balanced regime-context strata, and scene-content-disjoint splits. On this benchmark, we validate continuous bridge stops at non-dyadic Monte Carlo endpoints and show that the transport schedule can be reused as a physical prior that improves no-reference image quality across frozen diffusion and RF backbones.
\end{itemize}

\section{Related Work}

\paragraph{Rectified Flows and Continuous-Time Models}
Rectified Flows~\citep{liu2023rectified} and flow matching~\citep{lipman2023flow} learn a velocity field that transports samples from a source distribution to a target distribution along straight-line trajectories.
Diffusion models~\citep{Song2021ScoreSDE,Ho2020DDPM} and their refinements~\citep{Karras:2022:EDS} similarly evolve samples from noise to data through learned denoising or velocity fields.
These frameworks suggest a useful analogy to Monte Carlo integration: both move from high-variance states toward lower-variance structure.
Our setting differs in one key respect: rendering provides discrete real endpoints, not arbitrary synthetic interpolants.
We therefore train only on observed Monte Carlo endpoints, retaining arbitrary-NFE rollout while avoiding non-physical rung-interior supervision.

\paragraph{Monte Carlo Integration and Rendering}
Monte Carlo path tracing~\citep{Kajiya:1986:RTE} estimates light transport by averaging stochastic path-space samples governed by the rendering equation and physically based BRDFs~\citep{CookBRDF,burley2012physically}.
Classical rendering improves this estimator through sampling and variance-reduction strategies such as multiple importance sampling and Metropolis light transport~\citep{Veach:1995:MIS,Veach:1997:MLT}, as systematized in physically based rendering frameworks~\citep{Pharr:2023:PBR}.
Denoising methods---from edge-aware filters~\citep{tomasi1998bilateral} and learning-based Monte Carlo filters~\citep{Kalantari:2015:MLF} to KPCN~\citep{bako2017kernel}, OIDN~\citep{OIDN}, and temporal reconstruction~\citep{Chaitanya:2017:IRM}---typically operate on rendered images as noisy observations.
In contrast, we model the sequence of increasing-SPP render states itself as a continuous refinement process whose variance follows the $1/\sqrt{n}$ Monte Carlo law.

\paragraph{Neural Rendering with Generative Priors}
Recent work has explored injecting diffusion and flow-based generative priors into rendering pipelines.
Pretrained models support text-to-3D generation~\citep{poole2023dreamfusion}, novel view synthesis~\citep{liu2023zero123}, material estimation or editing~\citep{Sartor:2023:MGD,Kocsis:2023:IID,Sharma:2023:APC}, and can be steered through classifier-free guidance~\citep{Ho:2021:CFD} or ControlNet~\citep{Zhang:2023:ACC}.
Other work links diffusion models with rendering signals such as environment maps~\citep{zeng2024dilightnet,jin2024neural_gaffer} and G-buffers~\citep{zeng2024rgb,DiffusionRenderer}, while SDEdit~\citep{meng2021sdedit} refines noisy images through a generative prior.
These methods generally use the generator as a post-processing or conditioning module.
Our adapter instead injects a Monte Carlo transport direction derived from real render refinement into frozen diffusion and RF samplers as a physical prior.

\section{Method}

Monte Carlo rendering is a transport process.
The renderer, by accumulating samples, produces a natural refinement trajectory: samples flow from a noisy initial state toward the clean limit; the projection of this flow onto the image plane yields a continuous trajectory of images.
This section derives the governing SDE, defines the time coordinate that makes this trajectory representable as a continuous schedule, and formulates the transport field that operates on real render endpoints.

\begin{figure}[t]
  \centering
  \includegraphics[width=\linewidth]{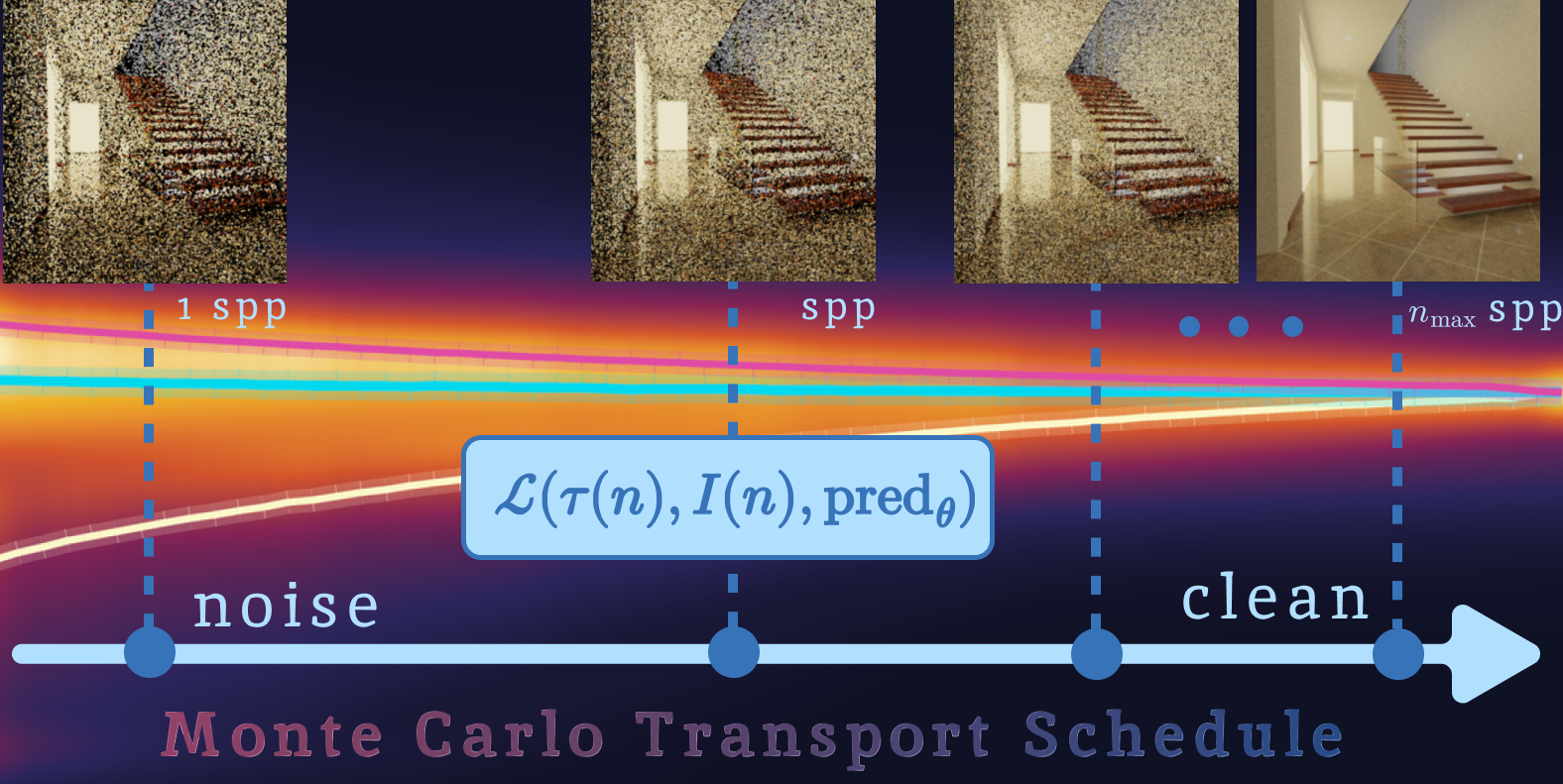}
  \caption{Monte Carlo Transport Scheduling. Top: a renderer produces a nested SPP ladder of real endpoint states, forming a natural refinement trajectory. Variance time $\tau(n)$, derived from the $1/\sqrt{n}$ convergence law, maps each SPP endpoint into a continuous solver clock spanning $\tau=0$ (noisy) to $\tau=1$ (clean). Bottom: the transport schedule is trained on real endpoints only, with two paired parameterizations---MC velocity for flow-based solvers and MC noise for diffusion-based denoisers---yielding a field that supports arbitrary-step rollout without retraining.}
  \Description{Overview of the Monte Carlo Transport Scheduling framework.}
  \label{fig:framework}
\end{figure}

\subsection{State: The Monte Carlo Transport SDE}

\paragraph{Projected particle distribution.}
A path tracer emits light paths $\gamma_i$ from the camera through the scene.
After $n$ independent samples, the renderer holds an empirical distribution of path particles:
\begin{equation}\label{eq:empirical}
  \mu_n = \frac{1}{n}\sum_{i=1}^{n}\delta_{\gamma_i}.
\end{equation}
Applying the measurement operator $\proj$, which maps each path to its radiance contribution $F(\gamma_i)$, produces the projected render state at $n$~spp:
\begin{equation}\label{eq:projection}
  I_n = \proj(\mu_n) = \frac{1}{n}\sum_{i=1}^{n} F(\gamma_i).
\end{equation}
As $n\to\infty$, the estimator converges to the ground truth $\mu = \E[F(\gamma)]$.

\paragraph{The Monte Carlo SDE}
The Central Limit Theorem characterizes the asymptotic distribution of $I_n$:
\begin{equation}\label{eq:clt}
  I_n \approx \mu + \frac{\sigma}{\sqrt{n}}\,\xi,\qquad \xi\sim\mathcal{N}(0,1),
\end{equation}
where $\sigma^2 = \Var[F(\gamma)]$.
The $1/\sqrt{n}$ scaling reveals the mechanism of noise decay: each additional sample narrows the estimator's distribution toward $\mu$ at a rate determined by the pixel's Monte Carlo variance.

To describe this as a continuous process, introduce a smooth mapping $n(\tau)$ with $n(\tau)\to\infty$ as $\tau\to 1$, and define $Y(\tau) \equiv I_{n(\tau)}$.
As new samples arrive over a small increment $\Delta\tau$, the estimator updates its value.
In the limit $\Delta\tau\to 0$, the update converges to the Monte Carlo SDE:
\begin{equation}\label{eq:mc_sde_general}
  dY(\tau)
  = \bigl(\mu - Y(\tau)\bigr)\,\frac{n'(\tau)}{n(\tau)}\,d\tau
    + \sigma\,\frac{\sqrt{|n'(\tau)|}}{n(\tau)}\,dW_\tau.
\end{equation}

The choice $n(\tau) = (1-\tau)^{-2}$ satisfies $n(0)=1$ and $\lim_{\tau\to 1}n(\tau)=\infty$, with a monotonic vanishing diffusion coefficient as the process approaches the clean limit.
Substituting gives the explicit form:
\begin{equation}\label{eq:mc_sde_explicit}
  dY(\tau)
  = \frac{2\bigl(\mu - Y(\tau)\bigr)}{1-\tau}\,d\tau
    + \sigma\sqrt{2(1-\tau)}\,dW_\tau.
\end{equation}

Equation~\eqref{eq:mc_sde_explicit} gives the continuous-time form of the Monte Carlo refinement process.
It expresses Monte Carlo integration as a continuous stochastic flow: a drift term pulls the process toward the true mean $\mu$, while a diffusion term, proportional to $\sqrt{1-\tau}$, injects noise that vanishes at $\tau=1$.
This provides a principled basis for treating sample-count-indexed render states as points along a continuous trajectory.

\paragraph{Variance-time embedding.}
The SDE reveals $1-\tau \propto 1/\sqrt{n}$ as the natural variance coordinate of Monte Carlo refinement, where the infinite-sample clean limit $\mu$ corresponds to $\tau=1$.
In practice, a renderer produces a finite SPP ladder $I_1, I_2, I_4, \dots, I_{n_{\max}}$; we take the highest-SPP endpoint $I_{n_{\max}}$ as the clean-limit estimate and assign it to $\tau=1$.
Normalizing the $1/\sqrt{n}$ relationship over this ladder defines the \emph{variance time}:
\begin{equation}\label{eq:tau_def}
  \tau(n) = \frac{\frac{1}{\sqrt{n_{\min}}} - \frac{1}{\sqrt{n}}}
                 {\frac{1}{\sqrt{n_{\min}}} - \frac{1}{\sqrt{n_{\max}}}}.
\end{equation}
Each SPP level now carries a time coordinate: $\tau=0$ at the noisiest state, $\tau=1$ at the cleanest.
The ladder becomes a continuous refinement trajectory $x_{\tau(n)} \equiv I_n$---real, physically valid states situated along the time axis of a Rectified Flow.

\subsection{Asymptotic Rendering Supervision}

The SPP ladder supplies real noisy-to-clean endpoints. We use the highest-SPP
render $I_{n_{\max}}$ as the clean estimate and learn each endpoint's relation
to it, expressed either in the coordinate of a diffusion denoiser or a flow
solver.

\paragraph{Noise and velocity targets.}
Our primary diffusion-style parameterization predicts the Monte Carlo noise
remaining in the current render endpoint:
\begin{equation}\label{eq:mc_noise}
  \epsilon^{\mathrm{MC}}_{\tau(n)} = I_n - I_{n_{\max}},\qquad n<n_{\max}.
\end{equation}
The output directly implies a clean endpoint
$\hat I_{\mathrm{clean}}=I_n-\epsilon_\theta(I_n,\tau(n),c)$, matching the
diffusion denoising pattern: predict noise, recover a clean estimate, and step
toward it. Our rendering variance time is clean-going, $\tau=0$ noisy and
$\tau=1$ clean; the usual diffusion noise time is $t_{\mathrm{diff}}=1-\tau$.

The RF-style parameterization uses the same clean chord but divides it by the
remaining variance time:
\begin{equation}\label{eq:final_clean}
  v^{\mathrm{MC}}_{\tau(n)}
  =
  \frac{I_{n_{\max}} - I_n}{1 - \tau(n)},\qquad n < n_{\max}.
\end{equation}
Thus MC velocity is the flow coordinate corresponding to the MC-noise target,
$v^{\mathrm{MC}}_{\tau(n)}=-\epsilon^{\mathrm{MC}}_{\tau(n)}/(1-\tau(n))$.
It matches the Rectified Flow view~\citep{liu2023rectified,lipman2023flow}, but
the supervision is anchored only on real Monte Carlo endpoints rather than
synthetic interpolants.

As a local ablation, the finite difference across two adjacent nested endpoints
with shared path prefix $\mu_n \subset \mu_{2n}$ defines
\begin{equation}\label{eq:local_vel}
  v^{\star}_{\tau(n)} = \frac{I_{2n} - I_n}{\tau(2n) - \tau(n)}.
\end{equation}
This local velocity is an adjacent-rung teacher, but its target is confined to
dyadic adjacency.

\paragraph{Asymptotic rendering training.}
Let $f_\theta$ be the predictor for the chosen parameterization. It receives a
real endpoint $I_k$, variance time $\tau_k$, and optional conditioning $c$.
Training optimizes
\begin{equation}\label{eq:loss}
  \mathcal{L}_{\mathrm{AR}}
  =
  \|f_\theta(I_k, \tau_k, c) - y^{\star}_k\|^2,
\end{equation}
where $y^\star_k$ is the MC-noise target in~\eqref{eq:mc_noise}, the MC-velocity
target in~\eqref{eq:final_clean}, or the local target in~\eqref{eq:local_vel}.
In each case, training samples only real ladder endpoints; non-physical
rung-interior states are never synthesized.

Both global parameterizations estimate the same clean endpoint:
$I_k-\epsilon_\theta(I_k,\tau_k,c)$ for MC noise, and
$I_k+(1-\tau_k)v_\theta(I_k,\tau_k,c)$ for MC velocity.

\paragraph{Arbitrary-NFE rollout.}
At inference, MC noise predicts a clean endpoint at the current state and the
solver takes a partial step toward it:
\begin{equation}\label{eq:noise_rollout}
  I_{\tau+\Delta\tau}
  =
  I_\tau+
  \frac{\Delta\tau}{1-\tau}
  \bigl(\hat I_{\mathrm{clean}}-I_\tau\bigr).
\end{equation}
The MC-velocity parameterization gives the corresponding RF-style Euler step:
\begin{equation}\label{eq:rollout}
  I_{\tau + \Delta\tau} = I_\tau + \Delta\tau \cdot v_\theta(I_\tau, \tau, c),
\end{equation}
arriving at the reconstructed clean state after $K$ steps in either coordinate.
Because the endpoint target encodes the mapping from any real state to the clean
limit, the same model can be evaluated with different numbers of solver steps
without retraining, yielding a native computation-fidelity tradeoff.
As shown in \Cref{sec:experiments}, this endpoint-anchored objective improves
NFE stability relative to synthetic-interpolation and purely local alternatives.
The same variance-time axis also connects rendering to continuous generative
samplers, whose trajectories evolve over a noise or velocity time.
We use this connection to define a compact transport adapter for frozen
generative backbones.

\subsection{Transport Adapter for Frozen Generative Backbones}
\label{sec:transport_adapter_method}

The Monte Carlo SDE also gives a principled target for reusing rendering noise
statistics inside pretrained iterative generators.
The key is to adapt the backbone in the correction space used by its sampler,
while deriving supervision from the same real Monte Carlo endpoints used above.
Let $B_\psi$ be a frozen RF or diffusion backbone, and let
$t_m=m(\tau)$ denote the model's time coordinate obtained by a fixed mapping from
variance time.
For a real endpoint $I_k$ at time $\tau_k$, we first map the frozen backbone
output into an adapter correction coordinate,
\begin{equation}
  u_{\mathrm{base}} =
  P\!\left(B_\psi(I_k,t_m), I_k, t_m\right),
\end{equation}
where $P$ is identity for an RF backbone. For a DDPM-style diffusion backbone,
$B_\psi$ predicts $\epsilon_\psi(I_k,t_m)$; we convert this prediction to its
implied residual/noise coordinate
\begin{equation}
  u_{\mathrm{base}} = I_k - \hat I_0,\qquad
  \hat I_0 =
  \frac{I_k-\sqrt{1-\bar\alpha_{t_m}}\,
  \epsilon_\psi(I_k,t_m)}
  {\sqrt{\bar\alpha_{t_m}}}.
\end{equation}
Thus $u_{\mathrm{base}}$ always denotes the frozen model's current estimate in
the coordinate that the adapter will bias.
We train a small residual adapter $A_\phi$ that only sees this frozen prediction
and the same model time:
\begin{equation}
  \Delta u_\phi = A_\phi(u_{\mathrm{base}}, t_m), \qquad
  \hat u = u_{\mathrm{base}} + \eta\,g(t_m)\Delta u_\phi ,
  \label{eq:adapter_inference}
\end{equation}
where $g$ is a late-stage injection schedule and $\eta$ controls the strength of
the physical bias.

The target $u^{\mathrm{MC}}$ depends on the backbone's prediction
parameterization but not on a new trajectory.
For an RF backbone, the sampler already operates in velocity space. The
corresponding Monte Carlo target is therefore the clean-going velocity implied
by the rendering SDE:
\begin{equation}
  u^{\mathrm{MC}}_{\mathrm{RF}}(I_k,\tau_k)
  =
  \frac{I_{n_{\max}} - I_k}{1-\tau_k}.
  \label{eq:adapter_rf_target}
\end{equation}
Thus the RF adapter learns a residual between two velocities: the Monte Carlo
velocity above and the frozen RF velocity prediction.
For a diffusion backbone, the corresponding coordinate is the Monte Carlo noise
remaining in the current endpoint,
\begin{equation}
  u^{\mathrm{MC}}_{\mathrm{diff}}(I_k,\tau_k)
  =
  I_k - I_{n_{\max}} .
  \label{eq:adapter_diff_target}
\end{equation}
This residual is the physical noise coordinate induced by the observed rendering
endpoint itself. Unlike a synthetic Gaussian diffusion process, the ladder does
not factor its uncertainty into a global schedule $\sigma_t$ times a standardized
unit-noise variable: the residual already carries the SPP-dependent magnitude
and the scene-, material-, visibility-, and pixel-dependent Monte Carlo
variance. The adapter therefore nudges the frozen model's native noise estimate
toward the Monte Carlo noise implied by the rendering process, without applying
an additional DDPM noise-scale normalization.
The adapter is therefore trained as a residual correction to the frozen
prediction,
\begin{equation}
  \Delta u^\star = u^{\mathrm{MC}} - u_{\mathrm{base}}, \qquad
  \mathcal{L}_{\mathrm{adapter}}
  =
  \rho\!\left(\Delta u_\phi-\Delta u^\star\right)
  + \lambda\|\Delta u_\phi\|^2 .
  \label{eq:adapter_loss}
\end{equation}
At generation time, $I_k$ is replaced by the sampler's current generated state.
For RF, the correction is added directly to the frozen velocity. For diffusion,
the correction is applied in residual space and then mapped back to the
backbone's epsilon parameterization before the scheduler step. In both cases,
the adapter remains a residual physical bias on the frozen prediction rather
than a replacement generator. We evaluate this reuse pathway in
\Cref{sec:generative_injection}.

\section{Experiments}
\label{sec:experiments}

We evaluate Monte Carlo Transport Scheduling along three axes: render refinement,
reuse as a physical prior for frozen generative samplers, and continuity over real
Monte Carlo states.

\begin{figure}[t]
  \centering
  \includegraphics[width=\columnwidth]{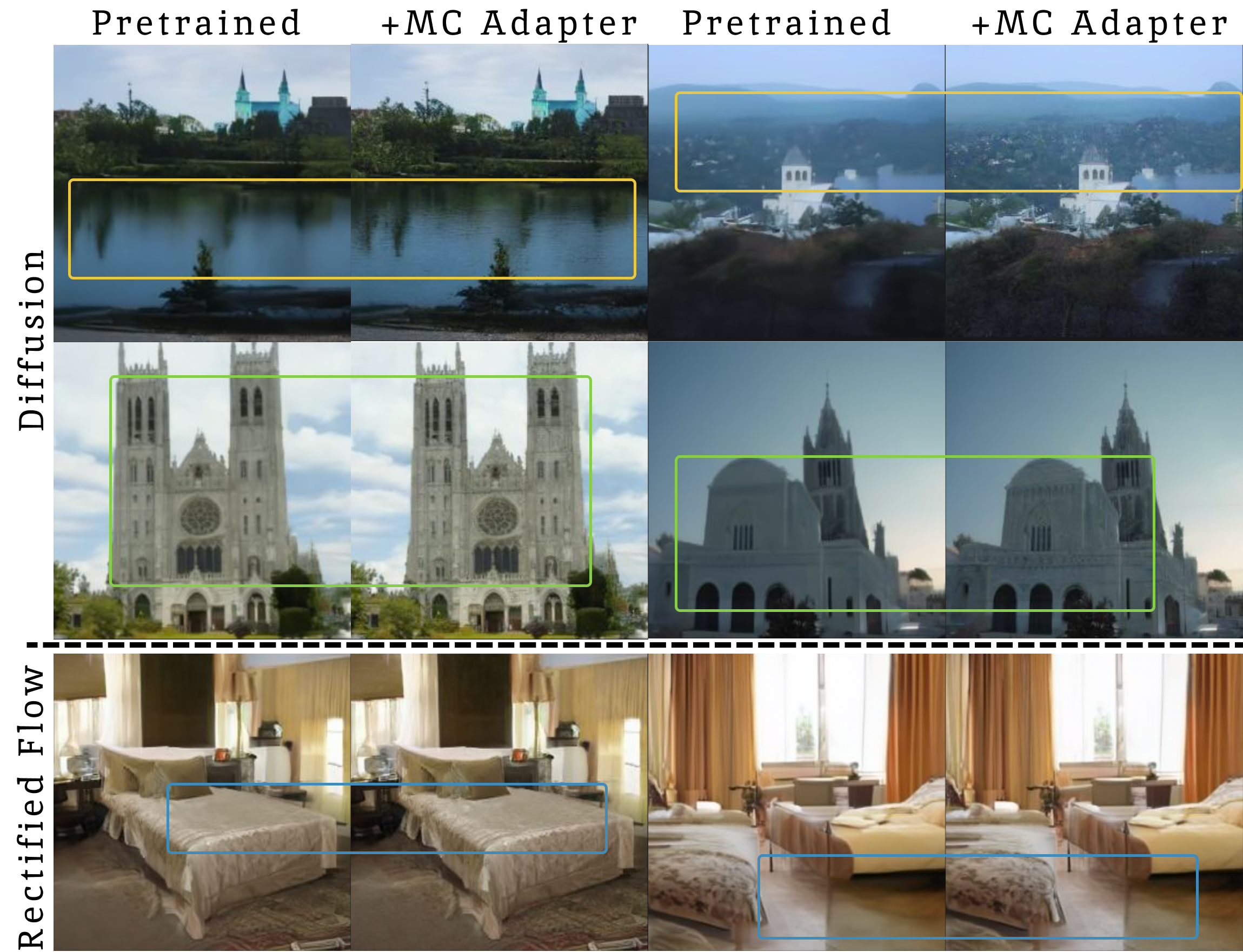}
  \caption{Generative trajectory injection. Same-seed samples from frozen
  pretrained backbones are paired with samples from the same backbones when the
  MC transport adapter is applied after $t=0.5$. Colored boxes highlight representative regions affected by the transport bias. Notably, upon introducing physical prior biases, the resulting images display structured details instead of uniform noise, clearly capturing elements like water ripples, church architecture, and bedding textures.}
  \label{fig:adapter}
  \Description{Comparison of frozen generative samples with and without
  MC transport-adapter trajectory injection for diffusion and Rectified Flow
  backbones.}
\end{figure}

\subsection{Dataset}
\label{sec:dataset}

\begin{table}[t]
\centering
\caption{Generative trajectory injection into frozen RF and DDPM samplers.
The adapter is trained on 128 samples and injected as a small residual
correction with $\eta=0.05$, leaving backbone weights fixed. Parenthesized
values are relative changes from the frozen backbone.}
\label{tab:gen_adapter_metrics}
\small
\resizebox{\columnwidth}{!}{%
\begin{tabular}{lrcccc}
\toprule
 Model & CLIPIQA$\uparrow$ & MUSIQ$\uparrow$ &
MANIQA$\uparrow$ & TOPIQ-NR$\uparrow$ \\
\midrule
RF Frozen & 0.266 & 51.222 & 0.376 & 0.500 \\
RF + transport adapter & 0.324 (+21.5\%) & 51.588 (+0.7\%) &
0.386 (+2.9\%) & 0.492 (-1.5\%) \\
DDPM Frozen & 0.493 & 52.204 & 0.430 & 0.532 \\
DDPM + transport adapter & 0.539 (+9.3\%) & 57.184 (+9.5\%) &
0.520 (+21.1\%) & 0.592 (+11.4\%) \\
\bottomrule
\end{tabular}%
}
\end{table}

\begin{figure*}[t]
  \centering
  \includegraphics[width=\textwidth]{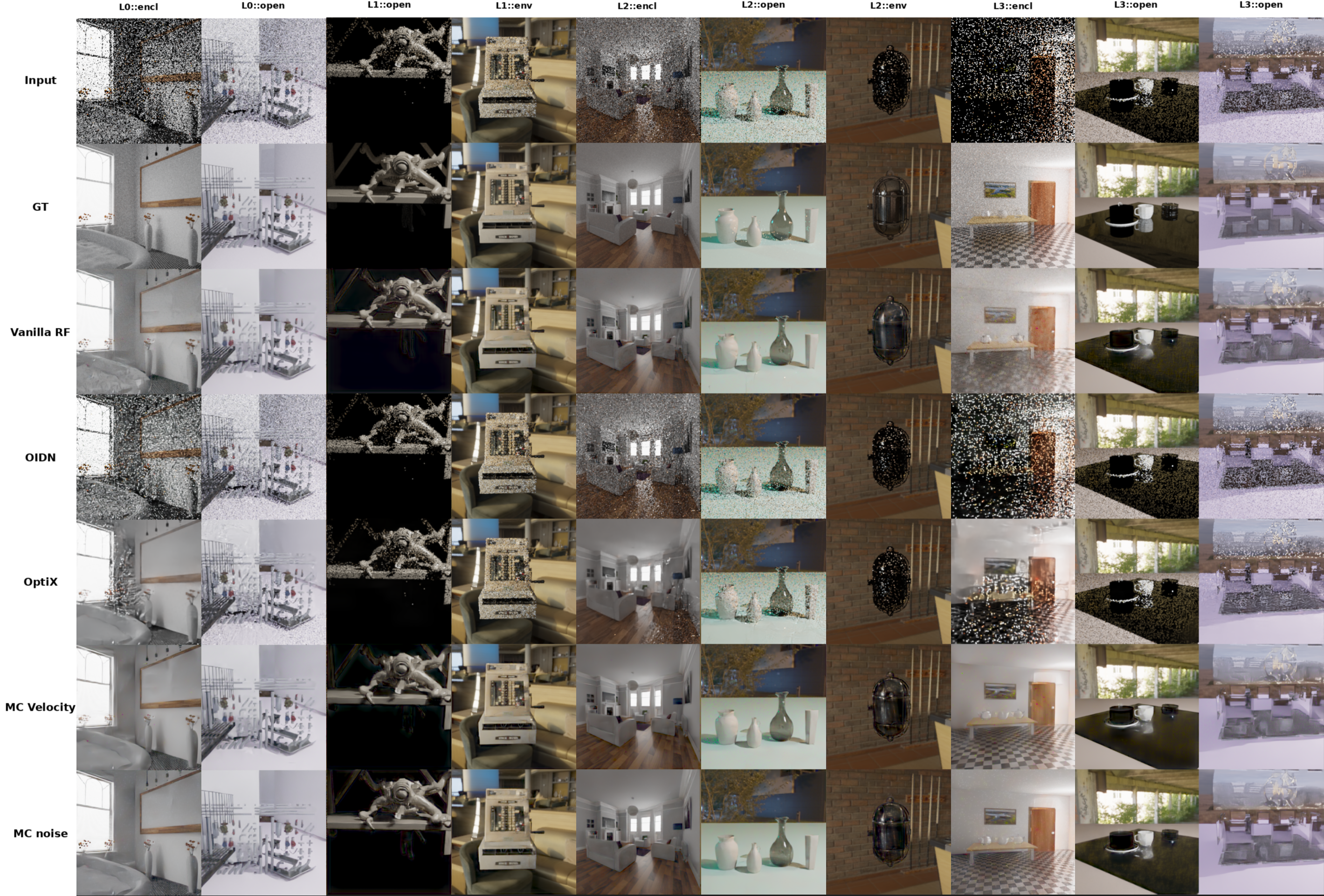}
  \caption{Stratified qualitative comparison on held-out scenes. Columns cover one
  sample from each of the 12 evaluation strata
  $\{\mathrm{L0,L1,L2,L3}\}\times\{\mathrm{enclosed,env,open}\}$. Rows show
  the 1~spp input, clean reference, and the four solver parameterizations used
  in \Cref{tab:main}. All learned methods use G-buffer input and NFE~9.
  Samples are selected from the held-out split to cover the
  transport-regime/context strata and expose visible method differences.
  The montage illustrates both the benchmark's transport-regime coverage and the
  structural differences that are under-penalized by preview-space scalar
  metrics.}
  \label{fig:visual_crop}
  \Description{A montage with one held-out scene from each rendering-regime and
  context stratum, comparing input, reference, Vanilla RF, local velocity,
  MC velocity, and MC noise outputs.}
\end{figure*}

Our training task is sensitive to the intermediate Monte Carlo states used for supervision: if the data do not contain physically valid refinement endpoints, a continuous solver can learn an artificial trajectory with poor rollout behavior.
We therefore construct a controlled Monte Carlo rendering benchmark whose scene variation, transport regimes, and nested sampling states are specified at generation time.

\paragraph{Data construction.}
We build a 30k-scene Mitsuba~\citep{mitsuba} pool from parametric layouts, Poly Haven HDRIs~\citep{PolyHaven}, and converted Objaverse objects~\citep{Deitke:2023:Objaverse}, with randomized materials, emitters, cameras, and exposure.
Scenes are balanced over 12 strata: four rendering regimes (L0 diffuse-only, L1 glossy, L2 refractive without caustics, L3 caustic-like) crossed with three transport contexts (environment-dominant, open multi-bounce, enclosed multi-bounce).
A 32~spp quality gate removes black or degenerate renders, leaving 23,883 usable samples.

\paragraph{Rendering and split protocol.}
Accepted scenes are rendered at $256\times256$ with a nested-prefix SPP ladder $1,2,4,\ldots,512$, so $I_{2k}$ contains the paths used by $I_k$ and all adjacent or final-clean targets are real Monte Carlo refinement states.
We use the 512~spp endpoint as both training reference and held-out reference.
Splits are disjoint by scene-content signature, and the main protocol trains on 5,000 samples and evaluates on 1,200 held-out samples, exactly 100 per stratum.
Metrics are linear HDR PSNR/MAE and preview-space PSNR/MAE, SSIM~\citep{wang2004image}, and LPIPS~\citep{zhang2018unreasonable}, using the same tone-map/gamma transform as saved qualitative PNGs.

\subsection{Continuous Solver Study}

\paragraph{Setup.}
We compare four solver parameterizations under identical architecture and data, following the protocol described above:
\emph{Vanilla RF}, a standard Rectified Flow baseline with single random-$t$ bridge supervision;
\emph{local velocity}, the adjacent-endpoint target in~\eqref{eq:local_vel};
\emph{MC velocity}, our endpoint-anchored flow parameterization with target $(I_{n_{\max}} - I_n) / (1 - \tau(n))$ and arbitrary NFE; and
\emph{MC noise}, the corresponding raw Monte Carlo noise parameterization.
Models receive either RGB-only input or G-buffer input (RGB + albedo + normal).
All solver tables and figures use the 60k checkpoints evaluated on the full
1,200-sample held-out split, unless a figure caption states otherwise. We
evaluate the same checkpoints at $N\in\{1,2,9,20,50\}$ model calls per image.
External denoiser references, OIDN and OptiX, use the RGB+G-buffer denoising
contract and therefore appear only as G-buffer-input reference levels without an
NFE axis.

\paragraph{Results.}
To summarize arbitrary-NFE behavior without privileging a single selected step
count, we integrate quality over the logarithmic compute axis. Let
$q_i=\mathrm{PSNR}(N_i)$ for $N_i\in\{1,2,9,20,50\}$ and
$s_i=\log N_i$. We report
$\mathrm{AUC}_{\log N}=(s_K-s_1)^{-1}\int q(s)\,ds$ using trapezoidal
interpolation, and
$\mathrm{TV}_{\log N}=(s_K-s_1)^{-1}\sum_i |q_{i+1}-q_i|$.

\Cref{tab:main} reports the solver formulation comparison in both linear HDR
and preview space. Across the log-NFE range, MC noise obtains the highest
AUC$_{\log N}$ in both spaces, while MC velocity gives the lowest
TV$_{\log N}$ in most settings and therefore a particularly steady rollout
curve. This supports the
central claim that supervision on real Monte Carlo endpoints yields solver
parameterizations that are both accurate and less sensitive to the selected
rollout budget than purely local alternatives.

\begin{figure*}[t]
  \centering
  \includegraphics[width=\textwidth]{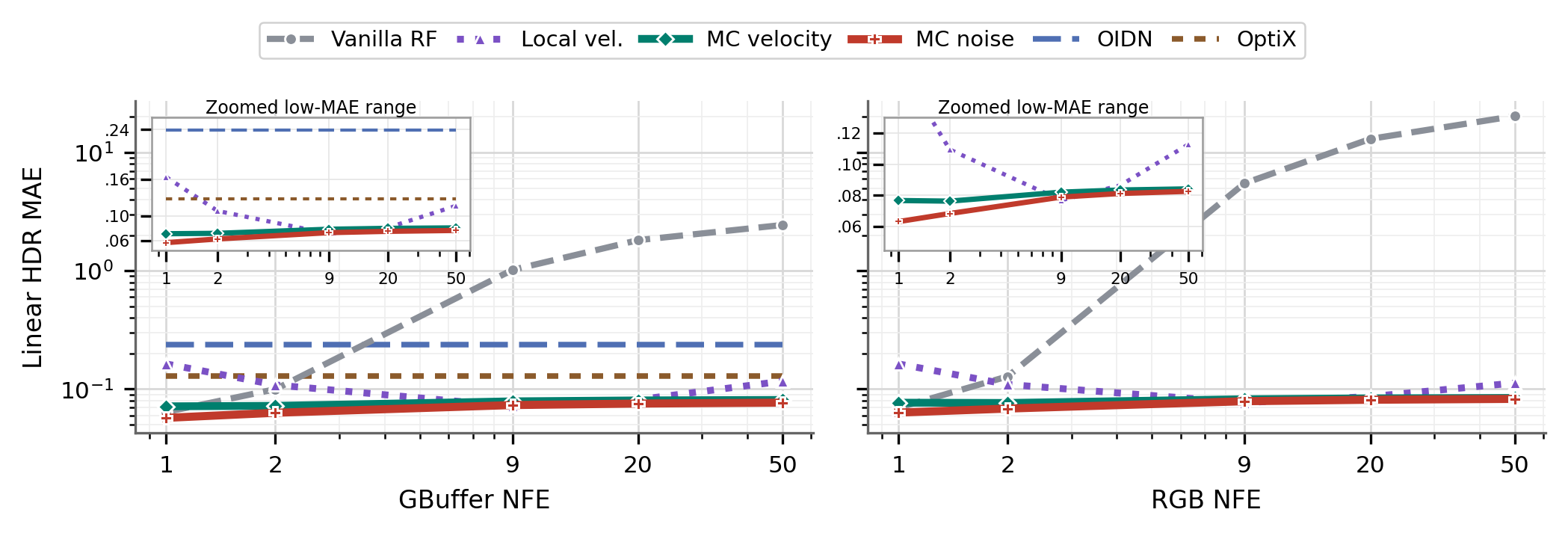}
  \caption{Linear HDR MAE across solver steps for G-buffer and RGB inputs. Unlike
  PSNR, which compresses rare extreme radiance failures through a logarithmic
  score, HDR MAE directly exposes large linear-radiance outliers. Vanilla RF
  develops large positive radiance outliers as NFE increases, while MC noise
  gives the lowest HDR error across most solver steps. OIDN and OptiX are
  G-buffer-input external denoisers and are shown as horizontal references only
  in the G-buffer panel. Insets zoom the low-error range.}
  \label{fig:hdr_mae_nfe}
  \Description{Two side-by-side line plots of linear HDR mean absolute error
  versus number of function evaluations for G-buffer and RGB inputs. Vanilla RF
  rises sharply with NFE, while MC noise and MC velocity remain in a low-error
  range; OIDN and OptiX appear as horizontal references in the G-buffer panel.}
\end{figure*}

\Cref{fig:hdr_mae_nfe} complements the PSNR results with a linear-radiance
error view. The HDR MAE curves show that endpoint-anchored training keeps
large radiance outliers bounded across rollout lengths, while Vanilla RF can
retain competitive preview-space scores despite unstable linear radiance values.

\paragraph{Qualitative comparison.}

\begin{figure*}[t]
  \centering
  \includegraphics[width=\textwidth]{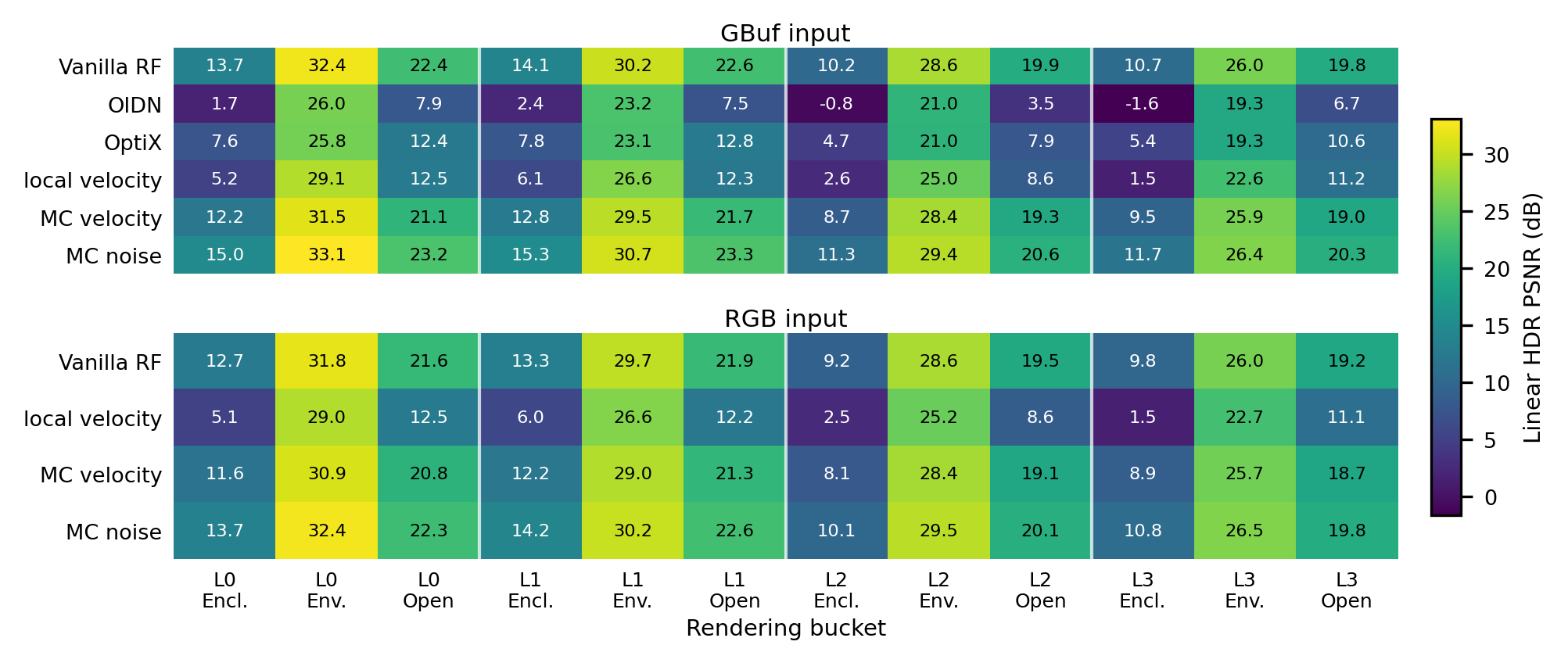}
  \caption{Bucket-wise linear HDR PSNR at NFE~1 on the full held-out
  evaluation. Solver rows follow \Cref{tab:main}; OIDN and OptiX are included
  as G-buffer-input external denoiser references. Columns cover the 12 balanced
  rendering strata, and each cell reports the mean PSNR over 100 held-out
  samples from that bucket.}
  \label{fig:bucket_psnr}
  \Description{Two heatmaps showing bucket-wise linear HDR PSNR for G-buffer
  and RGB inputs. The G-buffer panel includes Vanilla RF, OIDN, OptiX, local
  velocity, MC velocity, and MC noise; the RGB panel includes the four solver
  methods. Columns are the 12 rendering buckets formed by L0 through L3 regimes
  and enclosed, environment, and open transport contexts.}
\end{figure*}

\Cref{fig:visual_crop} provides a qualitative comparison.
The qualitative montage spans all regime-context groups in the evaluation split,
providing coverage beyond a single selected crop.
Vanilla RF often achieves competitive preview-space metrics, but the row reveals
structure drift, smoothing, and local transport failures that become especially
visible in glossy, refractive, and caustic-like strata.
The endpoint-anchored rows remove low-sample noise while more consistently
preserving scene layout, edges, and material boundaries, supporting the
quantitative gap in \Cref{tab:main} and the HDR-MAE failure mode shown in
\Cref{fig:hdr_mae_nfe}.
\Cref{fig:bucket_psnr} further visualizes the one-call HDR PSNR trend across
all 12 evaluation strata.

\subsection{Generative Model Trajectory Injection}
\label{sec:generative_injection}

We next test whether the transport adapter from
\Cref{sec:transport_adapter_method} can act as a physical bias for frozen
generative samplers.
We use two pretrained pixel-space backbones to cover both parameterizations in
our formulation: a DDPM-EMA church model~\citep{Ho2020DDPM} and an official
LSUN-bedroom Rectified Flow checkpoint~\citep{liu2023rectified}.
The backbones are never fine-tuned.
For each backbone, we train only the small adapter on the same 128-sample
balanced rendering subset used for the transport-supervision probe, using
preview-space SDR images and the variance-time mapping from
\Cref{sec:transport_adapter_method}.
The DDPM adapter is trained in the Monte Carlo noise coordinate, while the RF
adapter is trained in the corresponding Monte Carlo velocity coordinate.

At generation time, the adapter is injected only after the sampler reaches the
second half of its trajectory ($t_{\mathrm{start}}=0.5$), with strength
$\eta=0.05$ and a quadratic late-stage schedule; all backbone weights remain
fixed.
DDPM samples use 100 scheduler steps, and RF samples use 32 Euler steps.
We evaluate 100 paired same-seed generations for each backbone, comparing each
adapted sampler only against its own frozen baseline.
Metrics are no-reference image-quality scores---CLIPIQA~\citep{wang2023clipiqa},
MUSIQ~\citep{ke2021musiq}, MANIQA~\citep{yang2022maniqa}, and
TOPIQ-NR~\citep{chen2024topiq}---reported in their standard higher-is-better
direction.

\Cref{tab:gen_adapter_metrics} shows that the transport adapter improves all
reported DDPM metrics and three of four RF metrics, indicating that the learned
rendering direction can act as an effective physical prior for frozen generative
samplers.
The RF case is deliberately mild: the adapted samples remain close to the frozen
trajectory while improving CLIPIQA, MUSIQ, and MANIQA, with a small TOPIQ-NR
decrease.
The DDPM case shows a stronger response, with consistent gains across all four
reported no-reference metrics.
Qualitative paired comparisons are shown in \Cref{fig:adapter}.

\subsection{Continuous Rendering-State Bridge Validation}
\label{sec:bridge_validation}

\paragraph{Setup.}
The previous experiments evaluate Monte Carlo Transport Scheduling as an image refinement solver and as a
transport prior for frozen generative samplers.
We now test whether the learned noise predictor can induce a continuous bridge
between real Monte Carlo rendering states, rather than only predicting the final
clean endpoint.
Given a low-SPP endpoint $I_k$, MC noise first predicts the raw Monte Carlo
residual $r_\theta(I_k,\tau_k)$ and therefore an implied clean endpoint
$\hat I_\mathrm{clean}=I_k-r_\theta(I_k,\tau_k)$.
To stop at a later rendering endpoint $I_m$, we take the corresponding
variance-time fraction toward this clean estimate,
$\hat I_{k\rightarrow m}=I_k+
\frac{\tau_m-\tau_k}{1-\tau_k}(\hat I_\mathrm{clean}-I_k)$, and compare
$\hat I_{k\rightarrow m}$ against the real nested-prefix endpoint $I_m$.
This experiment uses RGB-only inputs to isolate the projected render-state
dynamics from auxiliary geometry conditioning.

\paragraph{Baselines.}
We compare three stopping rules.
\emph{Hold} directly reuses the source state $I_k$.
\emph{Local Step} is a discrete local-delta teacher rolled along the SPP
ladder.
\emph{MC noise} is our raw Monte Carlo noise predictor with the partial
clean-endpoint step above.
All methods use RGB-only inputs where applicable, and we report PSNR between the
predicted stop and the real target endpoint $I_m$.

\begin{table}[t]
\centering
\caption{Continuous rendering-state bridge validation on RGB-only Monte Carlo
endpoints. Values are average linear-RGB PSNR (dB) against the real
nested-prefix target endpoint $I_m$. The adjacent test averages 9 first-10-SPP
stops over 10 scenes; direct-to-10 averages 8 source endpoints over the same
scenes.}
\label{tab:bridge_stop_rgb}
\small
\resizebox{\columnwidth}{!}{%
\begin{tabular}{llcccc}
\toprule
Stop set & Target & Cases &
\multicolumn{3}{c}{PSNR to real target $\uparrow$} \\
\cmidrule(lr){4-6}
 & & & Hold & Local Step & MC noise \\
\midrule
Adjacent $1{\to}2,\ldots,9{\to}10$ & $I_2,\ldots,I_{10}$ & 90 &
22.160 & 21.006 & \textbf{23.084} \\
Direct $1$--$8{\to}10$~spp & $I_{10}$ & 80 &
17.410 & 19.074 & \textbf{20.809} \\
\bottomrule
\end{tabular}%
}
\end{table}

\begin{figure}[t]
  \centering
  \includegraphics[width=\columnwidth]{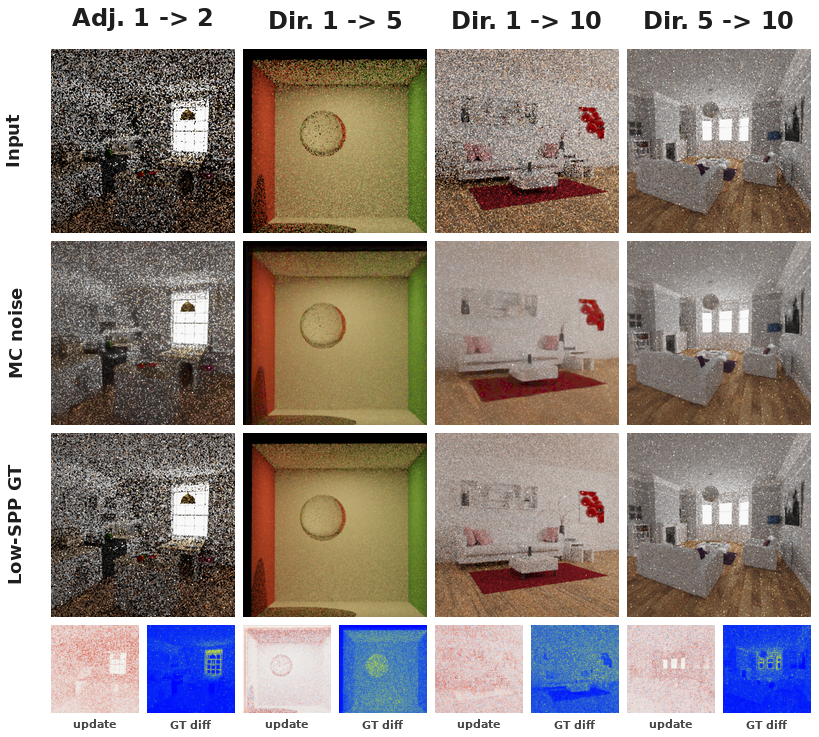}
  \caption{Continuous bridge stops on RGB-only Monte Carlo states.
  Each column starts from an input endpoint, estimates a clean endpoint by
  subtracting the predicted MC noise, and takes the partial variance-time step
  toward the requested later SPP endpoint.
  The predicted state is compared with the corresponding low-SPP ground truth.
  The bottom row shows the signed update direction and an absolute target-difference heatmap; heatmap colors are normalized per panel to localize residual structure.}
  \label{fig:bridge_stop_rgb}
  \Description{Visual comparison of input images, MC noise bridge stops, low-SPP ground truth endpoints, and delta maps for adjacent and direct rendering-state transitions.}
\end{figure}

\paragraph{Results.}
\Cref{tab:bridge_stop_rgb} reports the bridge-stop validation, and
\Cref{fig:bridge_stop_rgb} shows representative RGB-only bridge stops.
The \emph{Adjacent} test measures local continuity between consecutive early
endpoints, such as $I_5 \rightarrow I_6$.
The \emph{Direct to 10~spp} test measures whether different early states can be
stopped directly at the same non-dyadic endpoint $I_{10}$.
MC noise achieves the highest average PSNR in both groups.
The result indicates that the learned raw-noise predictor is not limited to
final-endpoint denoising.
Even from RGB-only projected rendering states, its implied clean endpoint
supports partial stops near later real Monte Carlo endpoints outside the
original dyadic ladder.
This supports interpreting Monte Carlo Transport Scheduling as a continuous bridge over projected rendering
states rather than only as a final-image mapping.
We treat this as a targeted continuity diagnostic rather than a replacement for
the full image-quality benchmark.

\section{Limitations}

Monte Carlo Transport Scheduling connects path-tracing refinement to continuous-time transport, and the current implementation keeps the auxiliary machinery minimal to center the core formulation.
Several extensions remain outside this minimal design.
First, we use G-buffer information only as optional albedo and normal channels; richer renderer-side signals such as timing statistics, filtered texture descriptors, or lighting features could strengthen the link between projected particle states and the rendering pipeline.
Second, although MC velocity yields stable arbitrary-NFE behavior, we do not explore progressive or trajectory distillation~\citep{salimans2022progressive,liu2023rectified} for further speed-quality tradeoffs.
Finally, production-scale generative models often operate in VAE latent spaces~\citep{rombach2022high,kingma2014autoencoding}, whose encoder variance can obscure the clean $1/\sqrt{n}$ SPP structure; adapting Monte Carlo transport to such latent spaces remains future work.

\section{Conclusion}

We have shown that Monte Carlo path tracing produces a natural refinement trajectory: the convergence of the estimator follows a continuous-time SDE, the time coordinate is variance time derived from the $1/\sqrt{n}$ law, and a transport field optimized over real rendered endpoints estimates the correction from noisy states toward the clean limit.
This transport framework defines Monte Carlo Transport Scheduling.
The schedule supports two paired parameterizations---MC velocity for flow-based solvers, MC noise for diffusion-based denoisers---and yields stable arbitrary-step inference with minimal degradation across the evaluated log-NFE range. The learned transport field is reusable: it serves as a continuous bridge between real rendering states at non-dyadic SPP endpoints, and as a physical prior that effectively links rendering-derived transport to frozen generative models.
Physically based rendering and generative modeling have long been treated as separate endeavors.
This paper shows that they can be connected through a shared view of continuous image transport.
The continuous formulation provides both practical and principled benefits: arbitrary-step stability without retraining, and a shared time axis that enables the transport field to act as a physical prior for frozen generative samplers.
By connecting them through a shared formulation, we provide a foundation on which future work can build: bringing richer renderer signals into the flow, resolving latent-space transport fidelity, and extending the transport prior to broader generative tasks.

\bibliographystyle{ACM-Reference-Format}
\bibliography{sample-base}

\begin{table*}[t]
\centering
\caption{Solver formulation comparison in linear HDR and RGB preview space. The
Space column marks linear HDR PSNR (HDR) or tone-mapped RGB preview PSNR
(Preview). GBuf denotes RGB input augmented with albedo and normal channels.
MC noise denotes the raw Monte Carlo noise parameterization evaluated from the
60k checkpoint. NFE is the number of model calls per image.
AUC$_{\log N}$ summarizes quality over the log-NFE
budget range; TV$_{\log N}$ is the normalized total PSNR variation over the
same range. Bold and underline denote best and second-best within each
metric/input block.}
\label{tab:main}
\small
\resizebox{\textwidth}{!}{%
\begin{tabular}{lllccccccc}
\toprule
Space & Method & Input & NFE1 & NFE2 & NFE9 & NFE20 & NFE50 & AUC$_{\log N}$ & TV$_{\log N}$ \\
\midrule
HDR & Vanilla RF & RGB & \underline{20.280} & \underline{20.395} & \underline{19.855} & \underline{19.583} & \underline{19.398} & \underline{19.931} & \underline{0.284} \\
 & local velocity & RGB & 13.581 & 17.098 & \textbf{20.099} & 19.082 & 16.552 & 18.040 & 2.573 \\
 & MC velocity & RGB & 19.560 & 19.594 & 19.179 & 19.060 & 18.994 & 19.282 & \textbf{0.162} \\
 & MC noise & RGB & \textbf{21.008} & \textbf{20.641} & 19.841 & \textbf{19.640} & \textbf{19.507} & \textbf{20.086} & 0.384 \\
\midrule
 & Vanilla RF & GBuf & \underline{20.878} & \underline{20.990} & 20.404 & \underline{20.124} & \underline{19.939} & \underline{20.495} & \underline{0.297} \\
 & local velocity & GBuf & 13.612 & 17.254 & \textbf{20.866} & 19.250 & 15.449 & 18.220 & 3.239 \\
 & MC velocity & GBuf & 19.963 & 19.975 & 19.603 & 19.509 & 19.459 & 19.702 & \textbf{0.135} \\
 & MC noise & GBuf & \textbf{21.687} & \textbf{21.265} & \underline{20.489} & \textbf{20.308} & \textbf{20.174} & \textbf{20.737} & 0.387 \\
\midrule
Preview & Vanilla RF & RGB & \underline{31.579} & \underline{31.792} & 31.558 & \underline{31.441} & \textbf{31.361} & \underline{31.577} & \textbf{0.165} \\
 & local velocity & RGB & 28.029 & 30.627 & \textbf{31.949} & 30.699 & 27.429 & 30.427 & 2.157 \\
 & MC velocity & RGB & 31.106 & 31.439 & 31.234 & 31.145 & 31.099 & 31.245 & \underline{0.172} \\
 & MC noise & RGB & \textbf{32.580} & \textbf{32.464} & \underline{31.730} & \textbf{31.459} & \underline{31.207} & \textbf{31.890} & 0.351 \\
\midrule
 & Vanilla RF & GBuf & \underline{32.134} & \underline{32.404} & \underline{32.189} & \textbf{32.072} & \textbf{31.993} & \underline{32.196} & \underline{0.174} \\
 & local velocity & GBuf & 28.052 & 30.738 & 32.058 & 30.001 & 25.935 & 30.165 & 2.589 \\
 & MC velocity & GBuf & 31.275 & 31.635 & 31.525 & 31.474 & 31.444 & 31.513 & \textbf{0.141} \\
 & MC noise & GBuf & \textbf{33.216} & \textbf{33.048} & \textbf{32.297} & \underline{32.010} & \underline{31.719} & \textbf{32.459} & 0.383 \\
\bottomrule
\end{tabular}%
}
\end{table*}

\begin{figure*}[t]
  \centering
  \includegraphics[width=\textwidth]{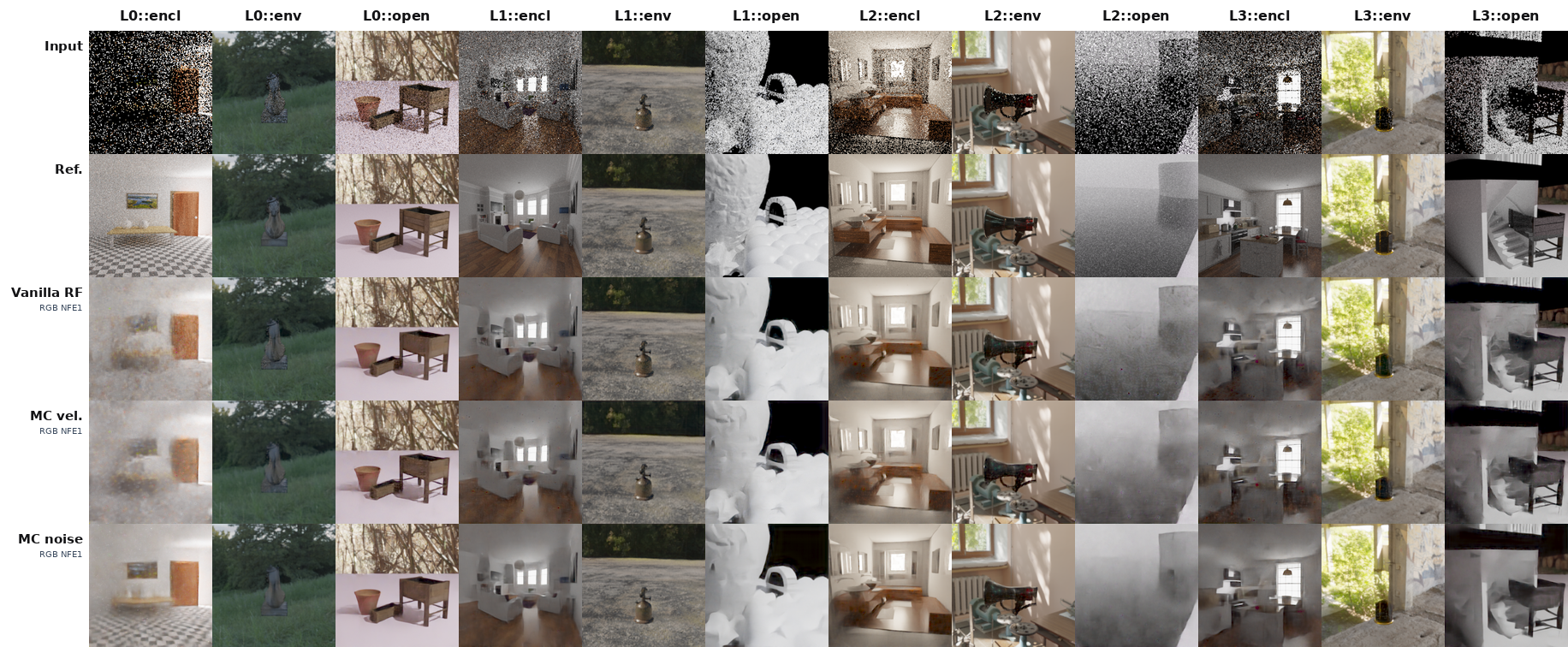}
  \caption{RGB-input qualitative comparison at NFE~1 across the 12 held-out
  rendering strata. Even with a single model call, MC noise preserves scene
  structure while maintaining strong visual quality.}
  \label{fig:bucket_rgb_nfe1}
  \Description{A qualitative montage over 12 rendering buckets comparing the
  noisy input, clean reference, Vanilla RF, MC velocity, and MC noise with RGB
  input at NFE 1.}
\end{figure*}

\end{document}